\documentclass[]{interact}

\usepackage{epstopdf}% To incorporate .eps illustrations using PDFLaTeX, etc.
\usepackage[caption=false]{subfig}% Support for small, `sub' figures and tables
\usepackage{natbib}% Citation support using natbib.sty
\bibpunct[, ]{(}{)}{;}{a}{}{,}% Citation support using natbib.sty
\renewcommand\bibfont{\fontsize{10}{12}\selectfont}% Bibliography support using natbib.sty
\usepackage{algorithmic}
\usepackage{algorithm}
\theoremstyle{plain}% Theorem-like structures provided by amsthm.sty

\theoremstyle{definition}

\theoremstyle{remark}

\begin{document}

%%%%%%%%%%%%%%%%%%%%%%%%%% Title %%%%%%%%%%%%%%%%%%%%%%%%% 
\title{ARST: Auto-Regressive Surgical Transformer for Phase Recognition from Laparoscopic Videos}

%%%%%%%%%%%%%%%%%%%%%%%%%% Authors %%%%%%%%%%%%%%%%%%%%%%%%%
%\author{
%\name{Xiaoyang Zou\textsuperscript{a}, Wenyong Liu\textsuperscript{b}, Junchen Wang\textsuperscript{c} Rong Tao\textsuperscript{a} and Guoyan Zheng\textsuperscript{a}\thanks{CONTACT Guoyan Zheng. Email: %guoyan.zheng@sjtu.edu.cn}}
%\affil{\textsuperscript{a} Institude of Medical Robotics, School of Biomedical Engineering, Shanghai Jiao Tong University, Shanghai, China; \textsuperscript{b} aaaa; \textsuperscript{c} bbbbb}
%}
\author{
\name{Xiaoyang Zou\textsuperscript{a}, Wenyong Liu\textsuperscript{b}, Junchen Wang\textsuperscript{c}, Rong Tao\textsuperscript{a} and Guoyan Zheng\textsuperscript{a}\thanks{Corresponding author: Prof. Dr. Guoyan Zheng, email: guoyan.zheng@sjtu.edu.cn}}
\affil{\textsuperscript{a}Institute of Medical Robotics, School of Biomedical Engineering, Shanghai Jiao Tong University, Shanghai 200240, China; \textsuperscript{b}Key Laboratory for Biomechanics and Mechanobiology of Ministry of Eduction, Beijing Advanced Innovation Center for Biomedical Engineering, School of Biological Science and Medical Engineering, Beihang University, Beijing 100083, China; \textsuperscript{c}School of Mechanical Engineering and Automation, Beihang University, Beijing 100091, China.}
}

\maketitle

%%%%%%%%%%%%%%%%%%%%%%%%% Abstract %%%%%%%%%%%%%%%%%%%%%%%%%
\begin{abstract}
Phase recognition plays an essential role for surgical workflow analysis in computer assisted intervention. Transformer, originally proposed for sequential data modeling in natural language processing, has been successfully applied to surgical phase recognition. Existing works based on transformer mainly focus on modeling attention dependency, without introducing auto-regression. In this work, an \textbf{A}uto-\textbf{R}egressive \textbf{S}urgical \textbf{T}ransformer, referred as ARST, is first proposed for on-line surgical phase recognition from laparoscopic videos, modeling the inter-phase correlation implicitly by conditional probability distribution. To reduce inference bias and to enhance phase consistency, we further develop a consistency constraint inference strategy based on auto-regression. We conduct comprehensive validations on a well-known public dataset Cholec80. Experimental results show that our method outperforms the state-of-the-art methods both quantitatively and qualitatively, and achieves an inference rate of 66 frames per second (fps).
\end{abstract}

%%%%%%%%%%%%%%%%%%%%%%%%% Keywords %%%%%%%%%%%%%%%%%%%%%%%%%

%%%%%%%%%%%%%%%%%%%%%%%%% Introduction %%%%%%%%%%%%%%%%%%%%%%%%%
\section{Introduction}
Surgical workflow analysis is an essential process in computer-assisted intervention (CAI) system, which is helpful for standardization and quality assessment of modern surgery\ \citep{maier2017surgical}. One of the crucial and challenging tasks is automatic surgical phase recognition. Accurate surgical phase recognition helps provide timely feedback and assistance for the surgeons during surgery, alert when abnormal cases occur, effectively improving the safety and intelligence level of modern operating room\ \citep{garrow2021machine}. In addition, analyzing surgical videos of skilled surgeons provides valuable training scheme for novice surgeons to develop their surgical skills\ \citep{zisimopoulos2018deepphase}.

Early works for surgical phase recognition are mainly based on the multi-dimensional state signals recorded under different operations\ \citep{padoy2012statistical, ahmadi2006recovery}. Compared with state signals, surgical videos are abundant and easier to acquire. However, recognizing surgical phases directly from surgical videos is a difficult task. For one thing, frames at different phases are highly similar in visual perception and the inter-phase correlation is quite challenging to model. For another, there always exist various of hard frames in surgery videos, caused by fast camera motion, produced gas, and camera out of surgical scene\ \citep{jin2017sv}.

% Related Works
With the development of deep learning, it has become the preferred technique for surgical phase recognition. Twinanda et al. first adopted convolutional neural network (CNN) to realize phase recognition purely based on visual frames\ \citep{twinanda2016endonet}. In order to model the temporal dependency, EndoLSTM\ \citep{twinanda2017vision} additionally utilized a long short-term memory (LSTM) network\ \citep{hochreiter1997long}. Jin et al. proposed an end-to-end network to integrate deep ResNet\ \citep{he2016deep} and LSTM networks\ \citep{jin2017sv}, and related multi-scale temporal information using non-local blocks\ \citep{jin2021temporal}. Czempiel et al. proposed TeCNO\ \citep{czempiel2020tecno}, which was a multi-stage causal temporal convolutional network (TCN)\ \citep{lea2016temporal, farha2019ms}, for temporal modeling based on ResNet features. Besides, post-processing methods such as prior knowledge inference\ \citep{jin2017sv} and scheme for hard frame detection and mapping\ \citep{yi2019hard} were developed.

Nowadays, transformer has demonstrated powerful capabilities for sequential data modeling in natural language processing\ \citep{vaswani2017attention}, and has been successfully applied to surgical workflow analysis. Czempiel et al. presented an attention-regularized transformer model based on self-attention\ \citep{czempiel2021opera}. Gao et al. designed a transformer to aggregate the spatial and temporal embeddings for better recognition performance\ \citep{gao2021trans}. Kondo realized surgical tool detection using transformer architecture\ \citep{kondo2021lapformer}. Nwoye et al. introduced new forms of spatial attention and semantic attention to recognize surgical action triplets\ \citep{nwoye2022rendezvous}. However, to the best of our knowledge, existing methods mostly focus on modeling the attention dependency, without exploiting the auto-regression characteristic of transformers.

In this work, we propose an \textbf{A}uto-\textbf{R}egressive \textbf{S}urgical \textbf{T}ransformer, referred as ARST, to accurately recognize surgical phases from laparoscopic videos. Our contributions can be summarized as following:

\begin{itemize}
\item We propose a novel transformer-based auto-regressive framework used for on-line surgical phase recognition. Phase prediction for each frame is conditionally dependent on predictions of previous frames, which can implicitly capture the inter-phase correlation. 

\item At inference stage, we propose a consistency constraint inference strategy, reducing inference bias while enhancing the consistency and reliability of auto-regressive phase predictions efficiently, independent of surgery-specific \textsl{a priori} information or post-processing methods. 

\item Comprehensive experiments are conducted to evaluate our method on the challenging Cholec80 dataset for surgical phase recognition. Our method shows superior capability for phase recognition, outperforming the state-of-the-art methods quantitatively and qualitatively.
\end{itemize}

%%%%%%%%%%%%%%%%%%%%%%%%% Related Works %%%%%%%%%%%%%%%%%%%%%%%%%
% \section{Related Works}

%%%%%%%%%%%%%%%%%%%%%%%%% Methodology %%%%%%%%%%%%%%%%%%%%%%%%%
\section{Methodology}

\begin{figure}
\centering
\resizebox*{13.2cm}{!}{\includegraphics{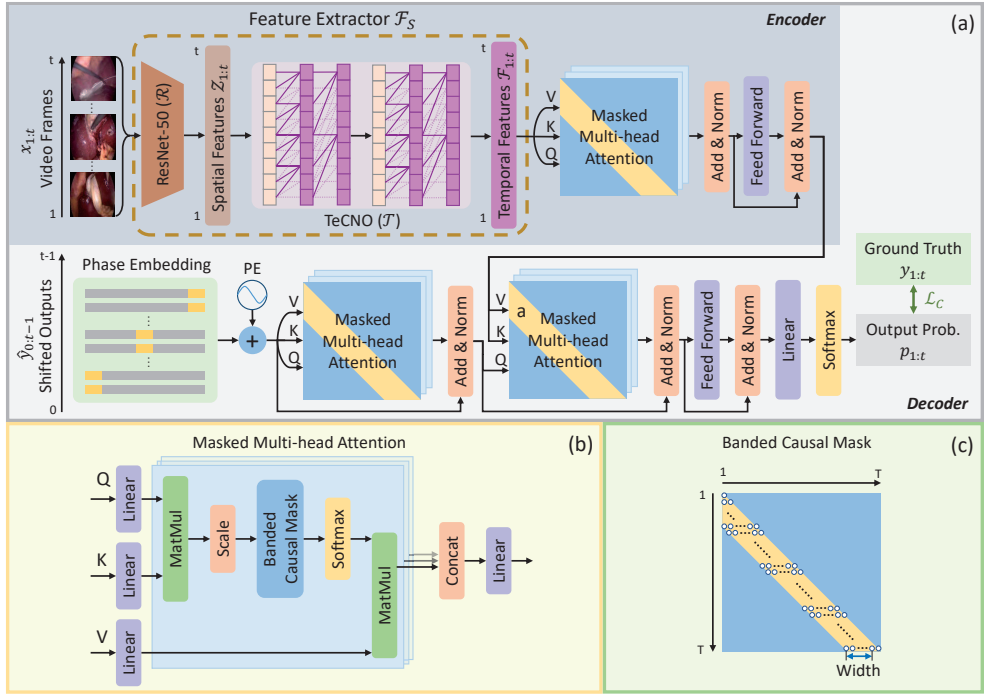}}
\hspace{0pt}
\caption{(a) A schematic overview of our proposed auto-regressive surgical transformer built on encoder-decoder architecture for on-line surgical phase recognition, (b) the masked multi-head attention with banded causal mask in detail, (c) illustration of the banded causal mask. PE: Positional Encoding.} 
\label{Methods-Pipeline}
\end{figure}

A schematic overview of our auto-regressive surgical transformer is presented in Figure~\ref{Methods-Pipeline}. The network consists of a frame-level feature extractor and a transformer-based encoder-decoder architecture for auto-regressive phase recognition.

\subsection{Feature Extractor}

Feature extractor is pre-trained to extract frame-level features from the video as the encoder's input embeddings. In our method, both spatial and temporal frame features are considered. Denote $\mathcal F_S$ as the feature extractor. Our spatial features are extracted by a deep residual CNN, the ResNet-50\ \citep{he2016deep}, denoted as $ \mathcal R$. Based on the spatial features, temporal features are extracted by a two-stage causal TCN, called TeCNO\ \citep{czempiel2020tecno}, denoted as $\mathcal T$. Assuming video length is $T$, for each video frame $x_t$ ($t\in[1,T]$), a frame feature $\mathcal F_t \in \mathbb{R}^{512}$ is extracted as presented below.

We first train a ResNet-50 is to extract spatial features, where each video frame is regarded as an independent image and frame-wise classification is conducted. Let $c$ denotes the number of phase classes. The network's input is the video frame $x_t \in \mathbb{R}^{H \times W \times C}$ and the output is a probability vector $p_t \in \mathbb{R}^c$. Specifically, following the global average pooling layer, the 2048-dimensional feature is first projected to a 512-dimensional feature, which is finally processed with a fully connected layer and a softmax layer to generate the probability vector. After training, for each input frame $x_t$, the 512-dimensional feature $\mathcal Z_t$, would serve as the spatial embeddings. It is used for further temporal feature extraction.

Based on the extracted spatial features, TeCNO\ \citep{czempiel2020tecno} is employed for efficient temporal modeling to extract temporal features at frame-level. Two-stage causal TCNs are cascaded. Each causal TCN contains 8 dilated causal convolutional layers, which can enlarge the receptive field and prevent the leakage of future information. Each convolutional layer has 512-dimensional feature maps. After training, for each input frame, the 512-dimensional feature $\mathcal F_t$, extracted at the second causal TCN, is used as the input to our auto-regressive surgical transformer.

%%%%%%%%%%%%%%%%%%%%%%%%% Auto-Regressive Surgical Transformer %%%%%%%%%%%%%%%%%%%%%%%%%
\subsection{Auto-Regressive Surgical Transformer}

Our auto-regressive surgical transformer, as shown in Figure~\ref{Methods-Pipeline}, is built on a lightweight one-layer encoder-decoder transformer\ \citep{vaswani2017attention}. The encoder consists two sub-layers, a masked multi-head attention layer and a fully connected feed-forward layer. Layer normalization\ \citep{ba2016layer} and residual connection\ \citep{he2016deep} are employed for each sub-layer. The decoder is designed similarly to the encoder, but has an additional masked multi-head attention layer to model the cross attention dependency between the encoder and decoder. For $t$ sequential frames starting from the first frame, frame features $\mathcal F_{1:t}$ serve as input to the encoder, while the shifted outputs $\hat y_{0:t-1}$ serve as input to the decoder. The final output probability is denoted as $p_{1:t}$, while the corresponding phase ground truth is denoted as $y_{1:t}$. Cross-entropy loss is calculated for model training.

Specific details of the masked multi-head attention layer with banded causal mask are also illustrated in Figure~\ref{Methods-Pipeline}. Three linear layers are employed for the projection of query (Q), key (K) and value (V) vectors with 8 heads. The embedding dimension of our model is 512 and the Q, K, V vectors of each head are in 64 dimensions (d). Scaled dot-product attention with a banded causal mask is computed for each head. Multiple heads enable the model to learn the attention dependencies from different aspects simultaneously. The outputs of all heads are concatenated together and projected to hidden dimension using a linear layer. 

Similar to human's mode of recognition, we believe that when predictions of previous frames are available, surgical phases can be recognized effectively within a limited range of visual frames. Long-term past information can be confusing and noisy for the decision making of current frame. In this regard, different from the usually adopted upper triangular mask\ \citep{vaswani2017attention}, we introduce a banded causal mask $M_{bc}$ in our work to limit the dependency range between frames while ensuring that future information is masked, thus supporting on-line phase recognition. Let $W$ denotes the width of banded causal mask. With element-wise multiplication, the region out of the $W$-width band is set as $-\infty$, resulting in null value for the softmax funtion. It means that only the previous $W$ frames before current frame are considered in modeling the attention dependency:

\begin{equation}
 Attention(Q,K,V) = Softmax(M_{bc}\circ \frac{Q K^T}{\sqrt{d}})V
\end{equation}

It's noteworthy that our phase inference is conducted sequentially in an auto-regressive fashion. When the phase of $t$-th frame is to be recognized, in addition to the frame features, all the predicted phases of previous frames $\hat{y}_{0:t-1}$ would serve as the input to transformer decoder, which can be regarded as conditional information. This can be expressed in terms of a conditional probability distribution $p(\hat{y}_t | \hat{y}_{0:t-1}, \mathcal F_{1:t})$. As the final prediction is determined by $p(\hat{y}_{1:T}|\mathcal F_{1:T})$, we can factorize it as a continued product of the conditional probability distribution according to Bayes' theorem. In this way, we can perform the auto-regressive prediction frame by frame to complete phase recognition of a whole video.

\vspace{-2mm}
\begin{equation}
 p(\hat{y}_{1:T}|\mathcal F_{1:T}) = \prod \limits_{t=1}^T p(\hat{y}_t | \hat{y}_{0:t-1}, \mathcal F_{1:t})
\end{equation}

At training stage, in order to speed up the training process and force the output as close as possible to the ground truth, we exploit the teacher forcing strategy\ \citep{williams1989learning} to parallelize the training process of our model. Instead of feeding shifted predicted outputs $\hat y_{0:t-1}$ each time for sequential phase generation, the whole shifted ground truth $y_{0:T-1}$ is used as the decoder's input to predict $\hat{y}_{1:T}$ for parallel training. During training, the inter-phase correlation can be implicitly captured by modeling the conditional probability distribution using attention layers, enabling the model to learn the pattern of phase transition actively.

One key component of designing our auto-regressive framework is to embed predicted phases as the input to the transformer decoder. Specifically, shifted outputs $\hat y_{0:t-1}$ need to be encoded as 512-dimensional phase embeddings $\mathcal E_{0:t-1}$. Similar to one-hot encoding, we first divide a 512-dimensional phase embedding vector into $c$ segments $S_{1:c}$ evenly according to the number of phase classes. We then assign binary value 1 to the $i$-th segment $S_i$ and keep other segments as 0 to obtain the encoding for the $i$-th phase ($i \in [1, c]$). This encoding strategy increases the distance between different phases in feature space compared with ordinary one-hot encoding.

\vspace{-2mm}
\begin{equation}
\mathcal E = {\begin{matrix} \underbrace{0,0,...,0} \\ S_1 \end{matrix}  \begin{matrix} ,......,\underbrace{1,1,...,1} ,......, \\ S_i \end{matrix} \begin{matrix} \underbrace{0,0,...,0} \\ S_c \end{matrix}}, ~~\hat y = i \in [1, c]
\end{equation}

However, without order information, identical phase embeddings at different positions in the video sequence can not be distinguished. In this regard, after phase embedding, positional encoding (PE) is added to describe the frame position in video of each phase through sine and cosine hybrid functions\ \citep{vaswani2017attention}. The dimension index is denoted as $i$.

\vspace{-2mm}
\begin{equation}
PE(t,2i) = sin(\frac{t}{10000^{2i/512}})
\end{equation}

\vspace{-5mm}
\begin{equation}
PE(t,2i+1) = cos(\frac{t}{10000^{2i/512}})
\end{equation}

\subsection{Consistency Constrained Inference}

Frequently jumped prediction is a serious problem in surgical phase recognition, which may have limited impact on the recognition performance, but lead to poor result in terms of consistency and reliability. Generally, jumped predictions are caused by low-quality frame features extracted from hard video frames, resulting in misrecognized phase transitions during on-line inference. Due to the nature of auto-regression, feeding incorrect predictions to the decoder will result in certain bias of the conditional probability distribution. The bias can be accumulated and thus has an negative impact on predictions of future frames. To minimize the impact, our model should carefully consider whether the recognized phase transitions are actually present. In this regard, consistency constraint inference (CCI) strategy is proposed to optimize our auto-regressive inference, which is explained below.

% from t-1 to t 
Assume that phase transition occurs from time $t-1$ to time $t$ (i.e., $ P_{t} \neq P_{t-1}$) during inference. Rather than directly feeding the newly predicted phase $P_t$ to the decoder for further inference, we continue predicting the next $n$ frames between $[t+1, t+n]$ by keeping feeding phase $P_{t-1}$ to decoder. Only if all of the predicted phases for the next $n$ frames are identical to phase $P_{t}$, we believe that an actual phase transition occurs. Otherwise, we consider the predicted phase transition is unreliable, which may be caused by noise or hard frames. In such a situation, prediction at $t$-th frame should be modified to phase $P_{t-1}$ for further inference. In this work, n is empirically set as 10. It is worth to note that our CCI strategy is integrated with our auto-reggression framework, as shown in Algorithm~\ref{alg:alg1}. Advantages of our newly proposed CCI strategy over previously introduced post-processing methods such as prior knowledge inference (PKI)\ \citep{jin2017sv} include: 1) our CCI strategy is part of our on-line inference procedure and thus is not a post-processing method; and 2) it does not use any surgery-specific \textsl{a priori} knowledge as in PKI\ \citep{jin2017sv} and thus can be extended to other types of workflow analysis tasks.

\begin{algorithm}[H]
\caption{Consistency Constraint Inference}
\label{alg:alg1}
\begin{algorithmic}
\STATE 
\STATE      {\textsc{INFERENCE}}     $(\mathbf{\mathcal F})$
\STATE      \hspace{0.5cm}      $\textbf{for}~t~\gets~1~\textbf{to}~T~\textbf{do}$
% \STATE      \hspace{1.0cm}      $\textbf{let}$ $\mathbf{P_a}$  $\textbf{denotes}$  $P_{t-1}$
\STATE      \hspace{1.0cm}        $ p_t~\gets~ARST(\mathcal F_{1:t},\hat{y}_{0:t-1})$
\STATE      \hspace{1.0cm}        $P_{t}~\gets~argmax(p_t)$
% \STATE      \hspace{1.0cm}      $\rm{/* phase~transition~happens */}$
% \STATE      \hspace{1.0cm}      $ if $  $ P_{t} \gets \mathbf{P_b} \neq \mathbf{P_a}$ 
\STATE      \hspace{1.0cm}      $ \textbf{if} $  $ P_{t} \neq P_{t-1}$ $ \textbf{then} $
\STATE      \hspace{1.5cm}      $\textbf{for}~j~\gets~1~\textbf{to}~n~\textbf{do}$
\STATE      \hspace{2.0cm}      $ \tilde{y}_{0:t+j-1}~\gets~\hat{y}_{0:t-1}~+~[\hat{y}_{t-1}]_{\times j}$
\vspace{0.5mm}
\STATE      \hspace{2.0cm}      $ \tilde{p}_{t+j}~\gets~ARST (\mathcal F_{1:t+j}, \tilde{y}_{0:t+j-1})$ 
\vspace{0.5mm}
\STATE      \hspace{2.0cm}      $ \tilde{P}_{t+j}~\gets~argmax(\tilde{p}_{t+j})$
\vspace{0.5mm}
\STATE      \hspace{2.0cm}      $\textbf{if}$ $\tilde{P}_{t+j} \neq P_{t}$ $\textbf{then}$
\vspace{0.5mm}
% \STATE      \hspace{2.0cm}      $P_{t} \gets \mathbf{P_b}$
% \STATE      \hspace{1.5cm}      $\textbf{else}$
\STATE      \hspace{2.5cm}      $P_{t} \gets P_{t-1}$
\STATE      \hspace{2.5cm}      $\textbf{Break}$

\end{algorithmic}
\label{alg1}
\end{algorithm}

%%%%%%%%%%%%%%%%%%%%%%%%% Experiments %%%%%%%%%%%%%%%%%%%%%%%%%
\section{Experiments}

%%%%%%%%%%%%%%%%%%%%%%%%% Data Description %%%%%%%%%%%%%%%%%%%%%%%%%
\subsection{Data Description}

\begin{table}
\tbl{Ids and descriptions of seven phases used in Cholec80.} 
{\begin{tabular}{ll} \toprule
ID & Phase descriptions                  \\ \midrule
P1 & Preparation               \\
P2 & Calot triangle dissection \\
P3 & Clipping and cutting      \\
P4 & Gallbladder dissection    \\
P5 & Gallbladder packaging     \\
P6 & Cleaning and coagulation  \\
P7 & Gallbladder retraction   \\ \bottomrule
\end{tabular}}
\label{PhaseDescription}
\end{table}

Our method is evaluated on a publicly available dataset Cholec80\ \citep{twinanda2016endonet}, which is a challenging laparoscopic video dataset of cholecystectomy surgery. The dataset contains 80 videos recorded at 25 frames per second (fps) with resolutions of either 1920$\times$1080 or 854$\times$480. Seven defined phases as shown in Table~\ref{PhaseDescription} are annotated manually at 25 fps, together with seven tool presence labels provided at 1 fps. Note that only the phase labels are used in our work. All the videos are sub-sampled to 1 fps for processing and training. We divide the dataset as 40 videos for training and the remaining 40 videos for testing. 8 videos in the training set are used for validation and hyper-parameters tuning.

%%%%%%%%%%%%%%%%%%%%%%%%% Evaluation Metrics %%%%%%%%%%%%%%%%%%%%%%%%%
\subsection{Evaluation Metrics}

To evaluate the performance of the proposed ARST quantitatively, four commonly used metrics are adopted, including accuracy, precision, recall and Jaccard index. For each video, the accuracy is evaluated by calculating the percentage of correctly recognized phases at video-level. Meanwhile, the precision, recall and Jaccard are first evaluated for each phase, and then averaged over all existing phases. We report the average value and standard deviation of each evaluation metric on the 40 testing cases.

%%%%%%%%%%%%%%%%%%%%%%%%% Training Protocol %%%%%%%%%%%%%%%%%%%%%%%%%
\subsection{Training Protocol}

Our feature extractor and ARST are implemented with PyTorch and trained on an NVIDIA TITAN RTX GPU. Spatial feature extractor ResNet-50 is initialized from a model pre-trained on ImageNet\ \citep{deng2009imagenet} and then trained for another 50 epochs using an SGD optimizer with 1e-4 learning rate. The resolution of input video frames is resized to 250$\times$250. Random 224$\times$224 cropping, flipping, rotating and color jittering are applied for data augmentation. For temporal modeling, we employ an Adam optimizer with 1e-4 learning rate to train the temporal feature extractor TeCNO for 50 epochs. After training, we fix feature extractors. We then train our ARST with Adam optimizer for 20 epochs. The learning rate is set as 1e-5. Each iteration uses frame features of a complete video, with batch size identical to the video's length. 

%%%%%%%%%%%%%%%%%%%%%%%%% Ablation Studies %%%%%%%%%%%%%%%%%%%%%%%%%
\subsection{Ablation Studies}

\begin{table}
\tbl{Ablative testing results for the banded causal mask with different width ($W$). Average values (\%) and standard deviations ($\pm$) are reported.}
{\begin{tabular}{lccccc} \toprule
$W$ & Accuracy & Precision & Recall & Jaccard\\ \midrule
0 & 84.83 $\pm$ 6.85 & 80.71 $\pm$ 8.93 & 79.92 $\pm$ 6.21 & 66.70 $\pm$ 9.61 \\
2 & 87.00 $\pm$ 6.87 & 83.64 $\pm$ 8.11 & 83.41 $\pm$ 5.79 & 71.28 $\pm$ 9.75  \\
5 & \textbf{87.62 $\pm$ 6.44} & \textbf{84.05 $\pm$ 6.67} & \textbf{84.04 $\pm$ 5.34} & \textbf{71.70 $\pm$ 8.97} \\
10 & 87.13 $\pm$ 6.71 & 82.66 $\pm$ 6.75 & 83.45 $\pm$ 5.86 & 70.51 $\pm$ 8.47  \\
20 & 86.10 $\pm$ 7.46 & 79.92 $\pm$ 9.02 & 80.93 $\pm$ 7.75 & 67.14 $\pm$ 10.48  \\
40 & 83.57 $\pm$ 7.80 & 74.52 $\pm$ 9.40 & 77.39 $\pm$ 9.76 & 62.50 $\pm$ 10.01 \\
60 & 82.03 $\pm$ 8.26 & 72.99 $\pm$ 8.83 & 78.40 $\pm$ 10.31 & 61.71 $\pm$ 9.85 \\
 \bottomrule
\end{tabular}}
\label{ablation-mask}
\end{table}

% Table 1
The width $W$ of the banded causal mask determines the range of previous frames considered in the attention dependency. We first evaluate the effect of the band width $W$ on the performance of our method. The ablation study results are reported in Table~\ref{ablation-mask}, the case with $W=0$ can be regarded as a degenerated attention. Our experimental results show that an appropriate $W$ is of great importance to achieve better performance. Larger $W$ allows to generate smoother predictions benefiting from rich temporal information, but phase transition will become more susceptible to noise or poor frame features. On the contrary, smaller $W$ allows limited temporal information, leading to frequently jumped phase predictions. From this table, one can see that setting $W=5$ achieves the best result.

% Table 2
Table~\ref{ablation-ar-cci} reports the quantitative evaluation of the ablative studies for auto-regression and the CCI strategy when either spatial features or temporal features are used as the input to our ARST encoder. Without auto-regression, we use phase predictions from ResNet-50\ \citep{he2016deep} or TeCNO\ \citep{czempiel2020tecno} as the input (after embedding) to our ARST decoder. No matter what types of features are used, the worst results are achieved when both auto-regression and the CCI strategy are not used. Adding auto-regression leads to better results. The improvement is substantial when either type of features are used, indicating that auto-regression is capable for temporal modeling and refinement. When both auto-regression and the CCI strategy are used, a performance boost of 4.4\% (when spatial features are used) or of 3.6\% (when temporal features are used) in terms of Jaccard is observed, demonstrating the superior performance of the proposed method for phase recognition.

\begin{figure}
\centering
\resizebox*{14cm}{!}{\includegraphics{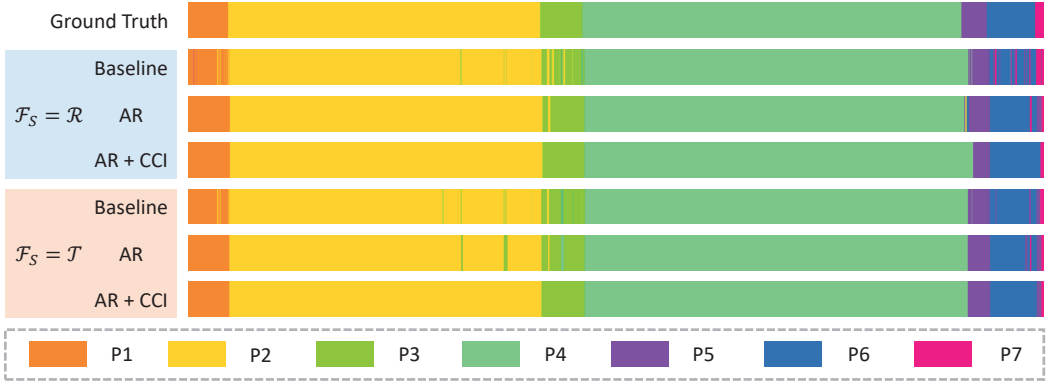}}
\hspace{5pt}
\caption{Qualitative illustration of the ablative testing results for auto-regression (AR) and consistency constrained inference (CCI) strategy. Two types of feature extractor $\mathcal F_S$, ResNet-50 ($\mathcal R$) and TeCNO ($\mathcal T$), are considered.} \label{figure-ablation-ar-cci}
\end{figure}

% Result Fig 1 
Moreover, visual comparisons (Figure~\ref{figure-ablation-ar-cci}) are provided to show the improvements qualitatively. Baseline is the case without using auto-regression and the CCI strategy. We can see that the predictions obtained when auto-regression is used are smoother than the baseline method, indicating that the model learns the pattern of phase transition well. Additionally, by incorporating our CCI strategy, the jumped predictions observed in the results obtained when auto-regression is used nearly disappear.

\begin{table}
\tbl{Ablative testing results for auto-regression (AR) and consistency constrained inference (CCI) strategy. Average metrics (\%) and the standard deviations ($\pm$) are reported. Two types of feature extractor $\mathcal F_S$, ResNet-50 ($\mathcal R$) and TeCNO ($\mathcal T$), are considered.}
{\begin{tabular}{llccccc} \toprule
$\mathcal F_S $ & AR & CCI & Accuracy & Precision & Recall & Jaccard\\ 
\midrule
$\mathcal R$ & - & - & 86.96 $\pm$ 7,47 & 84.24 $\pm$ 8.15 & 80.26 $\pm$ 7.88 & 68.81 $\pm$ 10.88  \\
$\mathcal R$ & \checkmark & - & 87.62 $\pm$ 6.44 & 84.05 $\pm$ 6.67 & 84.04 $\pm$ 5.34 & 71.70 $\pm$ 8.97  \\
$\mathcal R$ & \checkmark & \checkmark & \textbf{88.46 $\pm$ 6.81} & \textbf{84.93 $\pm$ 7.83} & \textbf{85.05 $\pm$ 7.24} & \textbf{73.16 $\pm$ 10.17} \\
\midrule
$\mathcal T$ & - & - & 88.19 $\pm$ 8.07 & 86.09 $\pm$ 7.19 & 83.31 $\pm$ 7.93 & 72.65 $\pm$ 10.93   \\
$\mathcal T$ & \checkmark & - & 88.23 $\pm$ 7.45 & 85.27 $\pm$ 7.50 & 84.94 $\pm$ 6.55 & 73.00 $\pm$ 10.25  \\
$\mathcal T$ & \checkmark & \checkmark & \textbf{89.27 $\pm$ 7.27} & \textbf{87.08 $\pm$ 7.25} & \textbf{86.82 $\pm$ 5.99} & \textbf{76.10 $\pm$ 9.62} \\ 
\bottomrule
\end{tabular}}
\label{ablation-ar-cci}
\end{table}

\begin{table}
\tbl{Quantitative comparison with baseline and state-of-the-art methods. Average values (\%) and standard deviations ($\pm$) are reported. Two types of feature extractor $\mathcal F_S$, ResNet-50 ($\mathcal R$) and TeCNO ($\mathcal T$), are considered.}
{\begin{tabular}{lccccc} \toprule
Methods & Accuracy & Precision & Recall & Jaccard \\ \midrule
ResNet-50\ \citep{he2016deep} & 81.54 $\pm$ 7.90 & 74.68 $\pm$ 8.87 & 75.29 $\pm$ 6.95 & 59.70 $\pm$ 9.58  \\
SV-RCNet\ \citep{jin2017sv} & 85.03 $\pm$ 6.81 & 79.70 $\pm$ 8.05 & 79.18 $\pm$ 6.26 & 65.46 $\pm$ 9.48  \\
TeCNO\ \citep{czempiel2020tecno} & 87.18 $\pm$ 7.72 & 82.66 $\pm$ 8.97 & 82.80 $\pm$ 6.46 & 69.62 $\pm$ 10.90  \\
Ours ($\mathcal F_S = \mathcal R $) & \textbf{88.46 $\pm$ 6.81} & \textbf{84.93 $\pm$ 7.83} & \textbf{85.05 $\pm$ 7.24} & \textbf{73.16 $\pm$ 10.17} \\
\midrule
Trans-SVNet\textsuperscript{*}\ \citep{gao2021trans} & 88.23 $\pm$ 7.97 & 84.98 $\pm$ 7.59 & 86.16 $\pm$ 6.40 & 73.83 $\pm$ 10.66 \\
Ours ($\mathcal F_S = \mathcal T$)\textsuperscript{*} & \textbf{89.27 $\pm$ 7.27} & \textbf{87.08 $\pm$ 7.25} & \textbf{86.82 $\pm$ 5.99} & \textbf{76.10 $\pm$ 9.62} \\
\bottomrule
\end{tabular}}
\tabnote{\textsuperscript{*} It denotes that a state-of-the-art temporal modeling method is involved, e.g. TeCNO.}
\label{comparison-sota}
\end{table}

%P1: Preparation; P2: Calot triangle dissection; P3: Clipping and cutting; P4: Gallbladder dissection; P5: Gallbladder packaging; P6: Cleaning and coagulation; P7: Gallbladder retraction.

%%%%%%%%%%%%%%%%%%%%%%%%% Comparison with the State-of-the-arts %%%%%%%%%%%%%%%%%%%%%%%%%

\begin{figure}
\centering
\resizebox*{14cm}{!}{\includegraphics{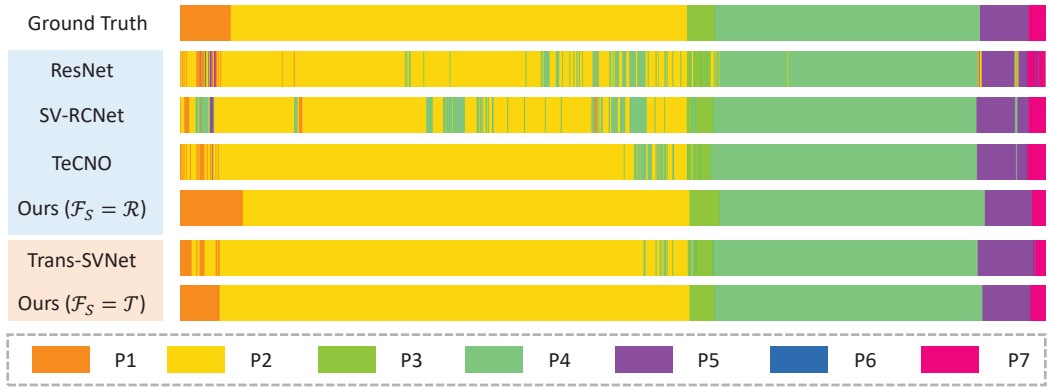}}
\hspace{5pt}
\caption{Qualitative comparison with baseline and state-of-the-art methods. Two types of feature extractor $\mathcal F_S$, ResNet-50 ($\mathcal R$) and TeCNO ($\mathcal T$), are considered.} 
\label{figure-comparison-sota}
\end{figure}

\subsection{Comparison with the State-of-the-arts}

% Table 3
We compare our method with several state-of-the-art methods. All the methods are trained without using any tool presence information. ResNet-50 is our backbone for spatial features extraction\ \citep{he2016deep}. SV-RCNet realizes end-to-end temporal modeling by seamlessly integrating ResNet and LSTM\ \citep{jin2017sv}. TeCNO is our backbone for temporal embeddings extraction, utilizing a two-stage causal TCN for temporal modeling\ \citep{czempiel2020tecno}. Trans-SVNet achieves superior recognition performance by aggregating spatial and temporal embeddings from ResNet and TeCNO\ \citep{gao2021trans}. 

In fairness, our method is compared with ResNet-50, SV-RCNet and TeCNO when spatial features are used. In addition, when temporal features are used, our method is only compared with Trans-SVNet, since both methods use temporal features extracted by TeCNO\ \citep{czempiel2020tecno}. All methods involved in the comparison are trained with same data augmentation and dataset partition setup. 

Table~\ref{comparison-sota} presents the quantitative evaluation results of the comparison. Please note that the results that we report here are based on the publicly available implementation of \ \citep{jin2017sv,czempiel2020tecno,gao2021trans} but evaluated on our own data split. From this table, one can see that, when spatial features are used, our method outperforms TeCNO by 1.3\%-3.5\% in all metrics and achieves a performance that is comparable to Trans-SVNet. Meanwhile, when temporal features are used, our method outperforms Trans-SVNet by 1\%-2.3\% in all metrics and achieves the best result among all compared methods. We additionally show a qualitative comparison in Figure~\ref{figure-comparison-sota}. No matter what types of features are used, the proposed method achieves smoother and more reliable predictions than other methods. Trans-SVNet achieves the second-best quantitative results but still produces frequently jumped predictions. In contrast, with the proposed method nearly no jumped prediction can be found. Benefiting from the limited band width $W$ of $M_{bc}$ and light-weighted network architecture, it takes on average 15.15 milliseconds for our method to process one frame, including the computing time for both feature extractors (spatial and temporal) and ARST. The proposed method can achieve an inference rate of 66 fps.

%%%%%%%%%%%%%%%%%%%%%%%%% Conclusion %%%%%%%%%%%%%%%%%%%%%%%%%
\section{Conclusion}

In this paper, we propose an auto-regressive surgical transformer, ARST, for accurate on-line phase recognition from laparoscopic videos. Phase prediction for each frame is conditionally dependent on predictions of previous frames, which can implicitly capture the inter-phase correlation. Besides, our novel consistency constraint inference strategy helps to reduce inference bias while enhancing the consistency and reliability of phase recognition. Our method outperforms state-of-the-art methods when evaluated on the public Cholec80 dataset, which may hold the potential to develop context-aware computer assisted interventional systems.

\section*{Acknowledgement}
The authors would like to thank the provider of Cholec80 dataset.

\section*{Disclosure statement}
The authors declare that they have no competing interests.

\section*{Funding}
This study was partially supported by Shanghai Municipal Science and Technology Commission via Project 20511105205, by the Natural Science Foundation of China via project U20A20199, and by Beijing Natural Science Foundation (Grant No.: L192056).

%%%%%%%%%%%%%%%%%%%%%%%%% Reference %%%%%%%%%%%%%%%%%%%%%%%%%

\bibliography{reference}
\bibliographystyle{tfcse}

\end{document}